\documentclass[10pt,twocolumn,letterpaper]{article}

\usepackage{cvpr}
\usepackage{times}
\usepackage{epsfig}
\usepackage{graphicx}
\usepackage{amsmath}
\usepackage{amssymb}
\usepackage{gensymb}

\usepackage{color}

\usepackage{booktabs}
\usepackage{multirow}
\usepackage{array}
\usepackage{xcolor}
\usepackage{subcaption}
\usepackage{graphicx}
\usepackage{hhline}
\usepackage{makecell}
\usepackage{colortbl}
\usepackage[normalem]{ulem}
\usepackage{soul}
\usepackage{url}
\definecolor{Gray}{rgb}{0.7,0.7,0.7}
\usepackage[square,comma,numbers,sort]{natbib}

\usepackage[
	pagebackref=true,
	breaklinks=true,
	letterpaper=true,
	colorlinks,
	bookmarks=false,
	citecolor=red,
	linkcolor=blue]{hyperref}

\newcolumntype{x}{>\small c}
\newcolumntype{L}[1]{>{\raggedright\let\newline\\\arraybackslash\hspace{0pt}}m{#1}}
\newcolumntype{C}[1]{>{\centering\let\newline\\\arraybackslash\hspace{0pt}}m{#1}}
\newcolumntype{R}[1]{>{\raggedleft\let\newline\\\arraybackslash\hspace{0pt}}m{#1}}

\newcolumntype{o}{>\small L}

\usepackage[pagebackref=true,breaklinks=true,letterpaper=true,colorlinks,bookmarks=false]{hyperref}

 \cvprfinalcopy 

\def\cvprPaperID{946} 

\ifcvprfinal\fi
\begin{document}

\title{A-Fast-RCNN: Hard Positive Generation via Adversary for Object Detection}

\author{Xiaolong Wang  \quad     \quad  Abhinav Shrivastava    \quad  \quad  Abhinav Gupta  \\
The Robotics Institute, Carnegie Mellon University \\
}

\maketitle

\begin{abstract}
How do we learn an object detector that is invariant to occlusions and deformations? Our current solution is to use a data-driven strategy -- collect large-scale datasets which have object instances under different conditions. The hope is that the final classifier can use these examples to learn invariances. But is it really possible to see all the occlusions in a dataset? We argue that like categories, occlusions and object deformations also follow a long-tail. Some occlusions and deformations are so rare that they  hardly happen; yet we want to learn a model invariant to such occurrences. In this paper, we propose an alternative solution. We propose to learn an adversarial network that generates examples with occlusions and deformations. The goal of the adversary is to generate examples that are difficult for the object detector to classify. In our framework both the original detector and adversary are learned in a  joint manner. Our experimental results indicate a 2.3\% mAP boost on VOC07 and a 2.6\% mAP boost on VOC2012 object detection challenge compared to the Fast-RCNN pipeline. We also release the  code\footnote{\url{https://github.com/xiaolonw/adversarial-frcnn}} for this paper.  
\end{abstract}

\vspace{-0.1in}
\section{Introduction}
\vspace{-0.05in}
The goal of object detection is to learn a visual model for concepts such as cars and use this model to localize these concepts in an image. This requires the ability to robustly model invariances to illuminations, deformations, occlusions and other intra-class variations. The standard paradigm to handle these invariances is to collect large-scale datasets which have object instances under different conditions. For example, the COCO dataset~\cite{coco} has more than 10K examples of cars under different occlusions and deformations. The hope is that these examples capture all possible variations of a visual concept and the classifier can then effectively model invariances to them. We believe this has been one of the prime reasons why ConvNets have been so successful at the task of object detection: they are able to use all this data to learn invariances.

\begin{figure}
    \centering
    \vspace{-0.25in}
    \includegraphics[width=\linewidth]{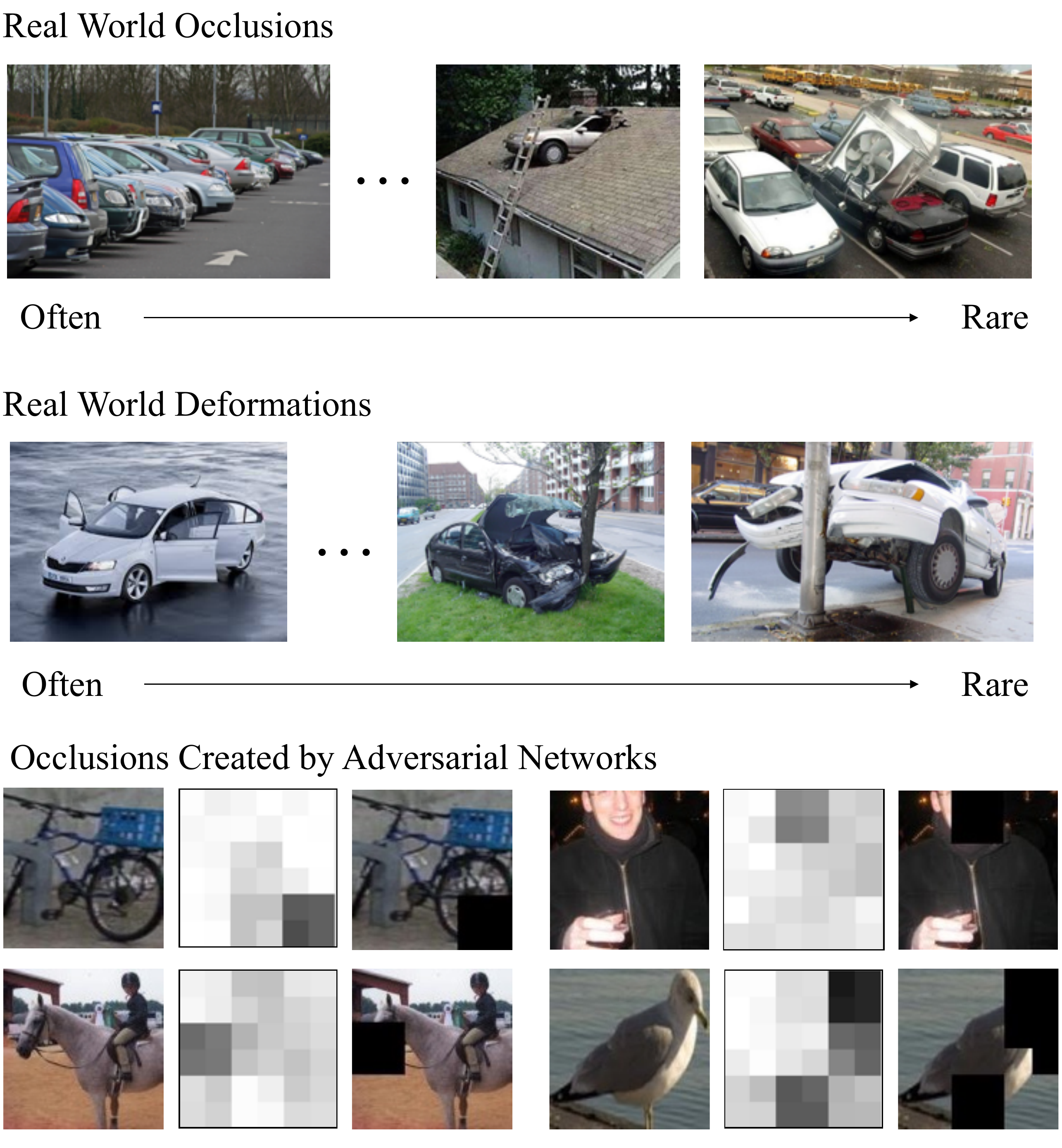}
    \caption{We argue that both occlusions and deformations follow a long-tail distribution. Some occlusions and deformations are rare. In this paper, we propose to use an adversarial network to generate examples with occlusions and deformations that will be hard for an object detector to classify. Our adversarial network adapts as the object detector becomes better and better. We show boost in detection accuracy with this adversarial learning strategy  empirically.}\label{fig:occlusions}
    \vspace{-0.2in}
\end{figure}

However, like object categories, we believe even occlusions and deformations follow a long-tail distribution. That is, some of the occlusions and deformations are so rare that there is a low chance that they will occur in large-scale datasets. 
For example, consider the occlusions shown in Figure~\ref{fig:occlusions}. We notice that some occlusions occur more often than others (\eg, occlusion from other cars in a parking garage is more frequent than from an air conditioner). Similarly, some deformations in animals are common (\eg, sitting/standing poses) while other deformations are very rare.

So, how can we learn invariances to such rare/uncommon occlusions and deformations? While collecting even larger datasets is one possible solution, it is not likely to scale due to the long-tail statistics.

Recently, there has been a lot of work on generating images (or pixels)~\cite{goodfellow2014generative,Denton15,Alec15}. One possible way to learn about these rare occurrences is to generate realistic images by sampling from the tail distribution. However, this is not a feasible solution since image generation would require training examples of these rare occurrences to begin with. Another solution is to generate all possible occlusions and deformations and train object detectors from them. However, since the space of deformations and occlusions is huge, this is not a scalable solution. It has been shown that using all examples is often not the optimal solution~\cite{shrivastavaOHEM,minibatchSVM} and selecting hard examples is better. Is there a way we can generate ``hard'' positive examples with different occlusions and deformations and without generating the pixels themselves?

How about training another network: an adversary that creates hard examples by blocking some feature maps spatially or creates spatial deformations by manipulating feature responses. This adversary will predict what it thinks will be hard for a detector like Fast-RCNN~\cite{frcn} and in turn the Fast-RCNN will adapt itself to learn to classify these adversarial examples. The key idea here is to create adversarial examples in convolutional feature space and not generate the pixels directly since the latter is a much harder problem. In our experiments, we show substantial improvements in performance of the adversarial Fast-RCNN (A-Fast-RCNN) compared to the standard Fast-RCNN pipeline.

\section{Related Work}
In recent years, significant gains have been made in the field of object detection. These recent successes build upon the powerful deep features~\cite{alex} learned from the task of ImageNet classification \cite{imagenet}. R-CNN \cite{rcnn} and OverFeat \cite{overfeat} object detection systems led this wave with impressive results on PASCAL VOC \cite{voc}; and in recent years, more computationally efficient versions have emerged that can efficiently train on larger datasets such as COCO~\cite{coco}. For example, Fast-RCNN~\cite{frcn} shares the convolutions across different region proposals to provide speed-up, Faster-RCNN ~\cite{renNIPS15fasterrcnn} and R-FCN~\cite{dai16rfcn} incorporate region proposal generation in the framework leading to a completely end-to-end version. Building on the sliding-window paradigm of the Overfeat detector, other computationally-efficient approaches have emerged such as YOLO~\cite{yolo16}, SSD~\cite{ssd16} and DenseBox~\cite{densebox}. Thorough comparisons among these methods are discussed in~\cite{Jonathan16}.

Recent research has focused on three principal directions on developing better object detection systems.
The first direction relies on changing the base architecture of these networks. The central idea is that using deeper networks should not only lead to improvements in classification~\cite{imagenet} but also object detection~\cite{voc,coco}. Some recent work in this direction include ResNet~\cite{resnet},  Inception-ResNet~\cite{inceptionresnet} and ResNetXt~\cite{Xie2016} for object detection.

The second area of research has been to use contextual reasoning, proxy tasks for reasoning and other top-down mechanisms for improving representations for object detection~\cite{ion,DeepMask,TDM17,FPN17,shrivastava16,Gidaris15,Zengeccv16}. For example,~\cite{shrivastava16} use segmentation as a way to contextually prime object detectors and provide feedback to initial layers. ~\cite{ion} uses skip-network architecture and uses features from multiple layers of representation in conjunction with contextual reasoning. Other approaches include using a top-down features for incorporating context and finer details~\cite{DeepMask,TDM17,FPN17} which leads to improved detections.

The third direction to improve a detection systems is to better exploit the data itself. It is often argued that the recent success of object detectors is a product of better visual representation and the availability of large-scale data for learning. Therefore, this third class of approaches try to explore how to better utilize data for improving the performance. One example is to incorporate hard example mining in an effective and efficient setup for training region-based ConvNets~\cite{shrivastavaOHEM}. Other examples of findind hard examples for training include ~\cite{simo2014fracking,wang2015unsupervised,loshchilov2015online}.

Our work follows this third direction of research where the focus is on leveraging data in a better manner. However, instead of trying to sift through the data to find hard examples, we try to generate examples which will be hard for Fast-RCNN to detect/classify. We restrict the space of new positive generation to adding occlusions and deformations to the current existing examples from the dataset. Specifically, we learn adversarial networks which try to predict occlusions and deformations that would lead to mis-classification by Fast-RCNN. Our work is therefore related to lot of recent work in adversarial learning~\cite{goodfellow2014generative,Denton15,Alec15,Mirza15,MathieuCL15,improvedGAN,pathakCVPR16context,CatGAN15,pix2pix2016}. 
For example, techniques have been proposed to improve adversarial learning for image generation~\cite{Alec15} as well as for training better image generative model~\cite{improvedGAN}.
\cite{improvedGAN} also highlights that the adversarial learning can improve image classification in a semi-supervised setting. However, the experiments in these works are conducted on data which has less complexity than object detection datasets, where image generation results are significantly inferior. Our work is also related to a recent work on adversarial training in robotics~\cite{pinto16}. However, instead of using an adversary for better supervision, we use the adversary to generate the hard examples.

\begin{figure*}
    \centering
    \includegraphics[width=1\textwidth]{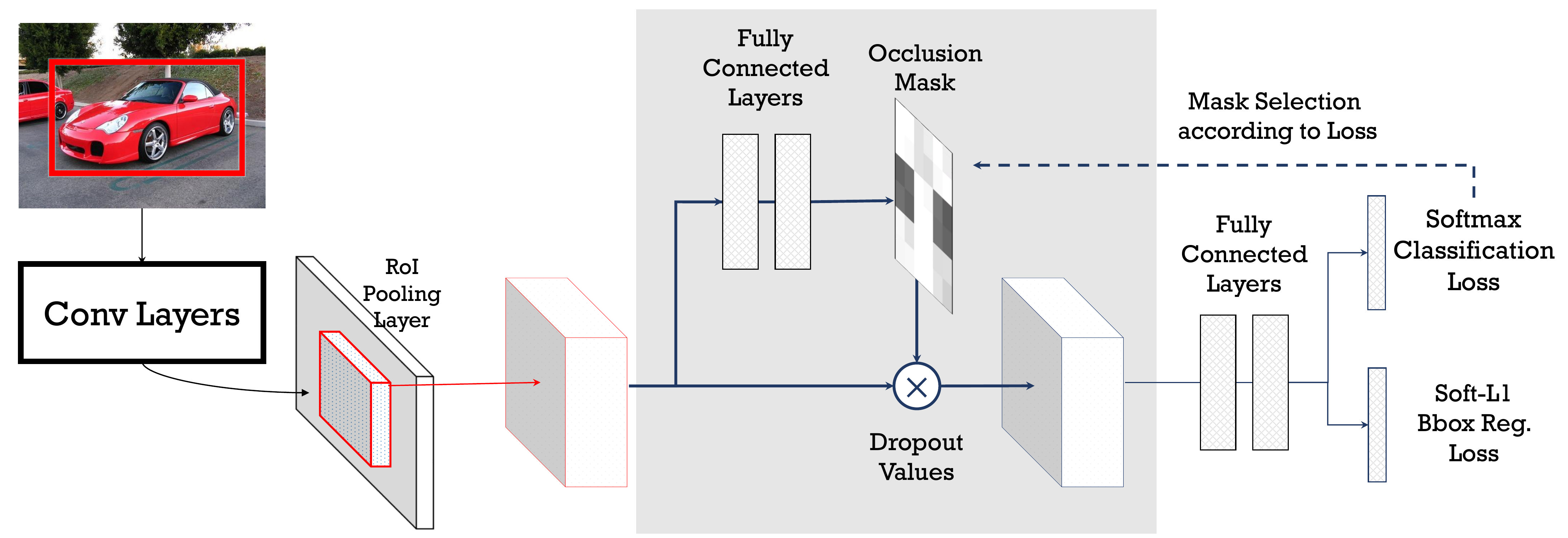}
    \caption{Our network architecture of ASDN and how it combines with Fast RCNN approach. Our ASDN network takes as input image patches with features extracted using RoI pooling layer. ASDN network than predicts an occlusion/dropout mask which is then used to drop the feature values and passed onto classification tower of Fast-RCNN.}\label{fig:network}
\end{figure*}

\section{Adversarial Learning for Object Detection}
Our goal is to learn an object detector that is robust to different conditions such as occlusion, deformation and illumination. We hypothesize that even in large-scale datasets, it is impossible to cover all potential occlusions and deformations. Instead of relying heavily on the dataset or sifting through data to find hard examples, we take an alternative approach. We actively generate examples which are hard for the object detector to recognize. However, instead of generating the data in the pixel space, we focus on a restricted space for generation: occlusion and deformation.

Mathematically, let us assume the original object detector network is represented as $\mathcal{F}(X)$ where $X$ is one of the object proposals. A detector gives two outputs $\mathcal{F}_c$ which represents class output and $\mathcal{F}_l$ represent predicted bounding box location. Let us assume that the ground-truth class for $X$ is $C$ with spatial location being $L$. Our original detector loss can be written down as, 
{\small
\begin{equation}
    \nonumber
    \mathcal{L}_\mathcal{F} = \mathcal{L}_{\text{softmax}}(\mathcal{F}_c(X), C) + \left[C \notin \text{bg}\right] \mathcal{L}_{\text{bbox}}(\mathcal{F}_l(X), L), 
\end{equation}
}
where the first term is the SoftMax loss and the second term is the loss based on predicted bounding box location and ground truth box location (foreground classes only). 

Let's assume the adversarial network is represented as $\mathcal{A}(X)$ which given a feature $X$ computed on image $I$, generates a new adversarial example. The loss function for the detector remains the same just that the mini-batch now includes fewer original and some adversarial examples. 

However, the adversarial network has to learn to predict the feature on which the detector would fail. We train this adversarial network via the following loss function,
{\small
\begin{equation}
    \nonumber
    \mathcal{L}_\mathcal{A} = - \mathcal{L}_{\text{softmax}}(\mathcal{F}_c(\mathcal{A}(X)), C).
\end{equation}
}
Therefore, if the feature generated by the adversarial network is easy for the detector to classify, we get a high loss for the adversarial network. On the other hand, if after adversarial feature generation it is difficult for the detector, we get a high loss for the detector and a low loss for the adversarial network.

\section{A-Fast-RCNN: Approach Details}
We now describe the details of our framework. We first give a brief overview of our base detector Fast-RCNN. This is followed by describing the space of adversarial generation. In particular, we focus on generating different types of occlusions and deformations in this paper. Finally, in Section~\ref{sec:exp}, we describe our experimental setup and show the results which indicate significant improvements over baselines.

\begin{figure*}
    \centering
    \includegraphics[width=1\textwidth]{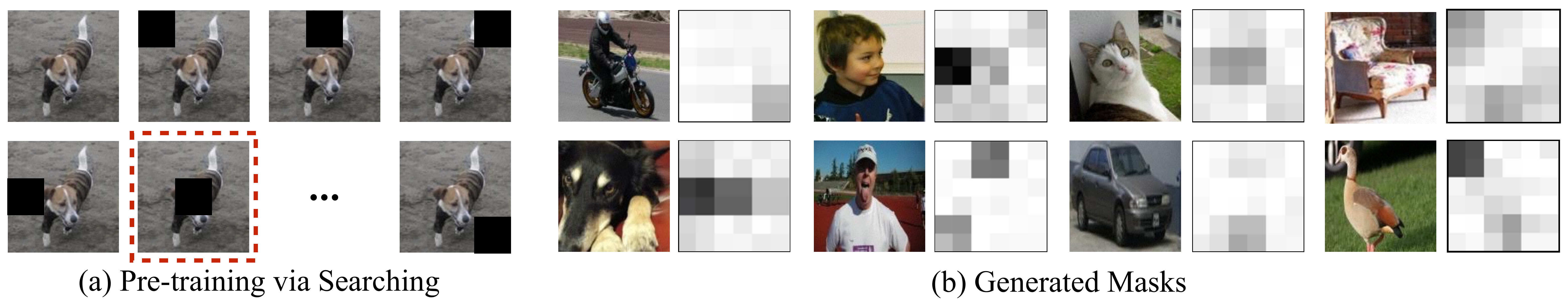}
    \caption{(a) Model pre-training: Examples of occlusions that are sifted to select the hard occlusions and used as ground-truth to train the ASDN network  (b) Examples of occlusion masks generated by ASDN network. The black regions are occluded when passed on to FRCN pipeline. }\label{fig:ASDN}
\end{figure*}

\subsection{Overview of Fast-RCNN}
We build upon the Fast-RCNN framework for object detection~\cite{frcn}. Fast-RCNN is composed of two parts: (i) a convolutional network for feature extraction; (ii) an RoI network with an  RoI-pooling layer and a few fully connected layers that output object classes and bounding boxes. 

Given an input image, the convolutional network of the Fast-RCNN takes the whole image as an input and produces convolutional feature maps as the output. Since the operations are mainly convolutions and max-pooling, the spatial dimensions of the output feature map will change according to the input image size. Given the feature map, the RoI-pooling layer is used to project the object proposals~\cite{Uijlings13} onto the feature space. The RoI-pooling layer crops and resizes to generate a fixed size feature vector for each object proposal. These feature vectors are then passed through fully connected layers. The outputs of the fully connected layers are: (i) probabilities for each object class including the background class; and (ii) bounding box coordinates. 

For training, the SoftMax loss and regression loss are applied on these two outputs respectively, and the gradients are back propagated  though all the layers to perform end-to-end learning.

\subsection{Adversarial Networks Design}
We consider two types of feature generations by adversarial networks competing against the Fast-RCNN (FRCN) detector. The first type of generation is occlusion. Here, we propose Adversarial Spatial Dropout Network (ASDN) which learns how to occlude a given object such that it becomes hard for FRCN to classify. The second type of generation we consider in this paper is deformation. In this case, we propose Adversarial Spatial Transformer Network (ASTN) which learns how to rotate ``parts'' of the objects and make them hard to recognize by the detector. By competing against these networks and overcoming the obstacles, the FRCN learns  to handle object occlusions and deformations in a robust manner. Note that both the proposed networks ASDN and ASTN are learned simultaneously in conjunction with the FRCN during training. Joint training prevents the detector from overfitting to the obstacles created by the fixed policies of generation. 

Instead of creating occlusions and deformations on the input images, we find that operating on the feature space is more efficient and effective. Thus, we design our adversarial networks to modify the features to make the object harder to recognize. Note that these two networks are only applied during training to improve the detector. We will first introduce the ASDN and ASTN individually and then combine them together in an unified framework.

\subsubsection{Adversarial Spatial Dropout for Occlusion}
We propose an Adversarial Spatial Dropout Network (ASDN) to create occlusions on the deep features for foreground objects. Recall that in the standard Fast-RCNN pipeline, we can obtain the convolutional features for each foreground object proposal after the RoI-pooling layer. We use these region-based features as the inputs for our adversarial network. Given the feature of an object, the ASDN will try to generate a mask indicating which parts of the feature to dropout (assigning zeros) so that the detector cannot recognize the object.

More specifically, given an object we extract the feature $X$ with size $d \times d \times c$, where $d$ is the spatial dimension and  $c$ represents the number of channels (e.g., $c=256, d=6$ in AlexNet). Given this feature, our ASDN will predict a mask $M$ with $d \times d$ values which are either 0 or 1 after thresholding. We visualize some of the masks before thresholding in Fig.~\ref{fig:ASDN}(b). We denote $M_{ij}$ as the value for the $i$th row and $j$th column of the mask. Similarly, $X_{ijk}$ represents the value in channel $k$ at location $i,j$ of the feature. If $M_{ij} = 1$, we drop out the values of all the channels in the corresponding spatial location of the feature map $X$, i.e., $X_{ijk} = 0, \forall k$. 

\textbf{Network Architecture.} We use the standard Fast-RCNN (FRCN) architecture. We initialize the network using pre-training from ImageNet~\cite{imagenet}.  The adversarial network shares the convolutional layers and RoI-pooling layer with FRCN and then uses its own separate fully connected layers. Note that we do not share the parameters in our ASDN with Fast-RCNN since we are optimizing two networks to do the exact opposite tasks. 

\textbf{Model Pre-training.} In our experiment, we find it important to pre-train the ASDN for creating occlusions before using it to improve Fast-RCNN. Motivated by the Faster RCNN detector~\cite{renNIPS15fasterrcnn}, we apply stage-wise training here. We first train our Fast-RCNN detector without ASDN for 10K iterations. As the detector now has a sense of the objects in the dataset, we train the ASDN model for creating the occlusions by fixing all the layers in the detector.

\begin{figure*}[t]
    \centering
    \includegraphics[width=1\textwidth]{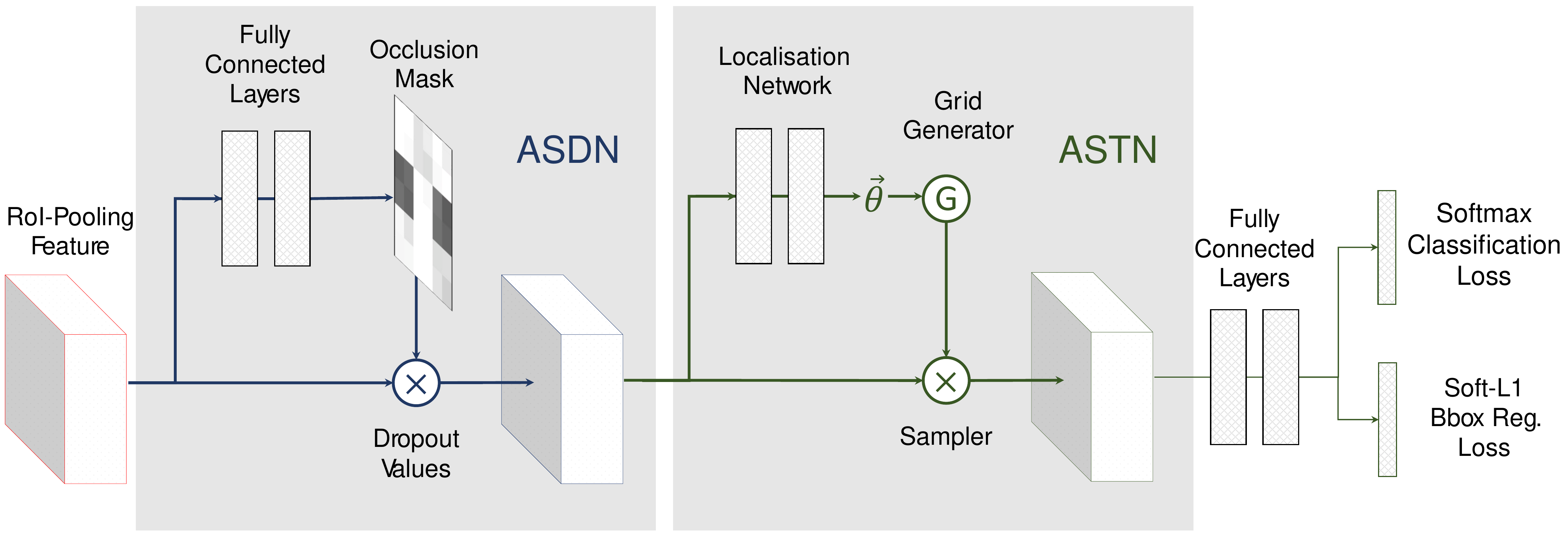}
    \caption{Network architecture for combining ASDN and ASTN network. First occlusion masks are created and then the channels are rotated to generate hard examples for training.}\label{fig:network_fusion}
\end{figure*}

\textbf{Initializing ASDN Network.} To initialize the ASDN network, given a feature map $X$ with spatial layout $d \times d$, we apply a sliding window with size $\frac{d}{3} \times \frac{d}{3}$ on it. We represent the sliding window process by projecting the window back to the image as~\ref{fig:ASDN}(a). For each sliding window, we drop out  the values in all channels whose spatial locations are covered by the window and generate a new feature vector for the region proposal. This feature vector is then passed through classification layers to compute the loss. Based on the loss of all the  $\frac{d}{3} \times \frac{d}{3}$ windows, we select the one with the highest loss. This window is then used to create a single $d \times d$ mask (with 1 for the window location and 0 for the other pixels). We generate these spatial masks for $n$ positive region proposals and obtain $n$ pairs of training examples $\{(X^{1}, \tilde{M}^{1}),..., (X^{n}, \tilde{M}^{n})\}$  for our adversarial dropout network. The idea is that the ASDN should learn to generate the masks which can give the detector network high losses. We apply the binary cross entropy loss in training the ASDN and it can be formulated as,
{\small
\begin{eqnarray}\label{eq:loss_mask}
\mathcal{L} =  - \frac{1}{n} \sum_{p}^{n} \sum_{i, j}^{d} [\tilde{M}_{ij}^p \mathcal{A}_{ij}(X^p) + (1 - \tilde{M}_{ij}^p) (1 - \mathcal{A}_{ij}(X^p)) ], 
\end{eqnarray}
}
where $\mathcal{A}_{ij}(X^p)$ represents the outputs of the ASDN in location $(i,j)$ given input feature map $X^p$. We train the ASDN with this loss for 10K iterations. We show that the network starts to recognize which part of the objects are significant for classification as shown in Fig.~\ref{fig:ASDN}(b). Also note that our output masks are different from the Attention Mask proposed in~\cite{ruslanattention}, where they use the attention mechanism to facilitate classification. In our case, we use the masks to occlude parts to make the classification harder.

\textbf{Thresholding by Sampling.} The output generated by ASDN network is not a binary mask but rather a continuous heatmap. Instead of using direct  thresholding, we use importance sampling to select the top $\frac{1}{3}$ pixels to mask out. Note that the sampling procedure incorporates stochasticity and diversity in samples during training. More specifically, given a heatmap, we first select the top $\frac{1}{2}$ pixels with top probabilities and randomly select $\frac{1}{3}$ pixels out of them to assign the value 1 and the rest of $\frac{2}{3}$ pixels are set to 0.

\textbf{Joint Learning.}  Given the pre-trained ASDN and Fast-RCNN model, we jointly optimize these two networks in each iteration of training. For training the Fast-RCNN detector, we first use the ASDN to generate the masks on the features after the RoI-pooling during forward propagation. We perform sampling to generate binary masks and use them to drop out the values in the features after the RoI-pooling layer. We then forward the modified features to calculate the loss and train the detector end-to-end. Note that although our features are modified, the labels remain the same. In this way, we create ``harder'' and more diverse examples for training the detector. 

For training the ASDN, since we apply the sampling strategy to convert the heatmap into a binary mask, which is not differentiable, we cannot directly back-prop the gradients from the classification loss. Alternatively, we take the inspirations  from the REINFORCE~\cite{reinforce} approach. We compute  which binary masks lead to significant drops in Fast-RCNN classification scores. We use only those hard example masks as ground-truth to train the adversarial network directly using the same loss as described in Eq.~\ref{eq:loss_mask}.

\vspace{-0.05in}
\subsubsection{Adversarial Spatial Transformer Network}
\vspace{-0.05in}
We now introduce the Adversarial Spatial Transformer Network (ASTN). The key idea is to create deformations on the object features and make object recognition by the detector difficult. Our network is built upon the Spatial Transformer Network (STN) proposed in~\cite{stn15}. In their work, the STN is proposed to deform the features to make classification easier. Our network, on the other hand, is doing the exact opposite task. By competing against our ASTN, we can train a better detector which is robust to deformations. 

\textbf{STN Overview.} The Spatial Transformer Network~\cite{stn15} has three components: localisation network, grid generator and sampler. Given the feature map as input, the localisation network will estimate the variables for deformations (e.g., rotation degree, translation distance and scaling factor). These variables will be used as inputs for the grid generator and sampler to operate on the feature map. The output is a deformed feature map. Note that we only need to learn the parameters in the localisation network. One of the key contribution of STN is making the whole process differentiable, so that the localisation network can be directly optimized for the classification objective via back propagation. Please refer to~\cite{stn15} for more technical details.

\textbf{Adversarial STN.} In our Adversarial Spatial Transformer Network, we focus on feature map rotations. That is, given a feature map after the RoI-pooling layer as input, our ASTN will learn to rotate the feature map to make it harder to recognize. Our localisation network is composed with 3 fully connected layers where the first two layers are initialized with fc6 and fc7 layers from ImageNet pre-trained network as in our Adversarial Spatial Dropout Network. 

We train the ASTN and the Fast-RCNN detector jointly. For training the detector, similar to the process in the ASDN, the features after RoI-pooling are first transformed by our ASTN and forwarded to the higher layers to compute the SoftMax loss. For training the ASTN, we optimize it so that the detector will classify the foreground objects as the background class. Different from training ASDN, since the spatial transformation is differentiable, we can directly use the classification loss to back-prop and finetune the parameters in the localisation network of ASTN. 

\textbf{Implementation Details.} In our experiments, we find it very important to limit the rotation degrees produced by the ASTN. Otherwise it is very easy to rotate the object upside down which is the hardest to recognize in most cases. We constrain the rotation degree within  10{\degree} clockwise and anti-clockwise. Instead of rotating all the feature map in the same direction, we divide the feature maps on the channel dimension into 4 blocks and estimate 4 different rotation angles for different blocks. Since each of the channel corresponds to activations of one type of feature, rotating channels separately corresponds to rotating parts of the object in different directions which leads to deformations. We also find that if we use one rotation angle for all feature maps, the ASTN will often predict the largest angle. By using 4 different angles instead of one, we increase the complexity of the task which prevents the network from predicting trivial deformations. 

\vspace{-0.1in}
\subsubsection{Adversarial Fusion}
\vspace{-0.05in}
The two adversarial networks ASDN and ASTN can also be combined and trained together in the same detection framework. Since these two networks offer different types of information. By competing against these two networks simultaneously, our detector become more robust. 

We combine these two networks into the Fast-RCNN framework in a sequential manner. As shown in Fig.~\ref{fig:network_fusion}, the feature maps extracted after the RoI-pooling are first forwarded to our ASDN which drop out some activations. The modified features are further deformed by the ASTN.

\section{Experiments}
\label{sec:exp}

We conduct our experiments on PASCAL VOC 2007, PASCAL VOC 2012~\cite{voc} and MS COCO~\cite{coco} datasets. As is standard practice, we perform most of the ablative studies on the PASCAL VOC 2007 dataset. We also report our numbers on the PASCAL VOC 2012 and COCO dataset. Finally, we perform a comparison between our method and the Online Hard Example Mining (OHEM)~\cite{shrivastavaOHEM} approach.

\begin{table*}[t]
\centering
\caption[caption]{\small {\bf VOC 2007 test} detection average precision (\%).  FRCN\raisebox{0.2ex}{$\star$} refers to FRCN \protect\cite{frcn} with our training schedule.}
\vspace{-0.1in}

\renewcommand{\arraystretch}{1.2}
\renewcommand{\tabcolsep}{1.2mm}
\resizebox{\linewidth}{!}{

\begin{tabular}{@{}L{2.1cm} !{\color{gray}\vrule}  L{1.2cm}   l r*{19}{x} @{}}
\Xhline{1pt}
method  & arch & mAP & aero      & bike      & bird      & boat      & bottle     & bus        & car        & cat        & chair      & cow        & table      & dog        & horse      & mbike      & persn     & plant      & sheep      & sofa       & train      & tv   \\
\Xhline{0.8pt}

FRCN~\cite{frcn}  &AlexNet  & 55.4 & 67.2 & 71.8 & 51.2 & 38.7 & 20.8 & 65.8 & 67.7 & 71.0 & 28.2 & 61.2 & 61.6 & 62.6 & 72.0 & 66.0 & 54.2 & 21.8 & 52.0 & 53.8 & 66.4 & 53.9 \\
FRCN\raisebox{0.2ex}{$\star$} &AlexNet  &57.0 & 67.3 & 72.1 & 54.0 & 38.3 & 24.4 & 65.7 & 70.7 & 66.9 & 32.4 & 60.2 & 63.2 & 62.5 & 72.4 & 67.6 & 59.2 & 24.1 & 53.0 & 60.6 & 64.0 & 61.5 \\
Ours (ASTN)  &AlexNet  & 58.1  & 68.7 & 73.4 & 53.9 & 36.9 & 26.5 & 69.4 & 71.8 & 68.7 & 33.0 & 60.6 & 64.0 & 60.9 & 76.5 & 70.6 & 60.9 & 25.2 & 55.2 & 56.9 & 68.3 & 59.9 \\
Ours (ASDN)  &AlexNet  & 58.5 & 67.1 & 72.0 & 53.4 & 36.4 & 25.3 & 68.5 & 71.8 & 70.0 & 34.7 & 63.1 & 64.5 & 64.3 & 75.5 & 70.0 & 61.5 & 26.8 & 55.3 & 58.2 & 70.5 & 60.5 \\
Ours (full)  &AlexNet  & 58.9 & 67.6 & 74.8 & 53.8 & 38.2 & 25.2 & 69.1 & 72.4 & 68.8 & 34.5 & 63.0 & 66.2 & 63.6 & 75.0 & 70.8 & 61.6 & 26.9 & 55.7 & 57.8 & 71.7 & 60.6 \\

\Xhline{0.5pt}
FRCN~\cite{frcn}  &VGG & 66.9 & 74.5 & 78.3 & 69.2 & 53.2 & 36.6 & 77.3 & 78.2 & 82.0 & 40.7 & 72.7 & 67.9 & 79.6 & 79.2 & 73.0 & 69.0 & 30.1 & 65.4 & 70.2 & 75.8 & 65.8 \\
FRCN\raisebox{0.2ex}{$\star$} &VGG  & 69.1 & 75.4 & 80.8 & 67.3 & 59.9 & 37.6 & 81.9 & 80.0 & 84.5 & 50.0 & 77.1 & 68.2 & 81.0 & 82.5 & 74.3 & 69.9 & 28.4 & 71.1 & 70.2 & 75.8 & 66.6  \\
Ours (ASTN)  &VGG  & 69.9 & 73.7 & 81.5 & 66.0 & 53.1 & 45.2 & 82.2 & 79.3 & 82.7 & 53.1 & 75.8 & 72.3 & 81.8 & 81.6 & 75.6 & 72.6 & 36.6 & 66.3 & 69.2 & 76.6 & 72.7  \\
Ours (ASDN)  &VGG  & 71.0 & 74.4 & 81.3 & 67.6 & 57.0 & 46.6 & 81.0 & 79.3 & 86.0 & 52.9 & 75.9 & 73.7 & 82.6 & 83.2 & 77.7 & 72.7 & 37.4 & 66.3 & 71.2 & 78.2 & 74.3 \\
Ours (full)  &VGG  & 71.4 & 75.7 & 83.6 & 68.4 & 58.0 & 44.7 & 81.9 & 80.4 & 86.3 & 53.7 & 76.1 & 72.5 & 82.6 & 83.9 & 77.1 & 73.1 & 38.1 & 70.0 & 69.7 & 78.8 & 73.1 \\

\Xhline{0.5pt}
FRCN\raisebox{0.2ex}{$\star$} &ResNet  & 71.8 & 78.7 & 82.2 & 71.8 & 55.1 & 41.7 & 79.5 & 80.8 & 88.5 & 53.4 & 81.8 & 72.1 & 87.6 & 85.2 & 80.0 & 72.0 & 35.5 & 71.6 & 75.8 & 78.3 & 64.3  \\
Ours (full)  &ResNet  & 73.6 &  75.4 & 83.8 & 75.1 & 61.3 & 44.8 & 81.9 & 81.1 & 87.9 & 57.9 & 81.2 & 72.5 & 87.6 & 85.2 & 80.3 & 74.7 & 44.3 & 72.2 & 76.7 & 76.9 & 71.4 \\

\Xhline{1pt}

\end{tabular}
}
\vspace{-0.05in}
\label{tab:voc2007}
\end{table*}

\begin{table*}[t]
\centering
\caption[caption]{\small {\bf VOC 2007 test} detection average precision (\%). Ablative analysis on the Adversarial Spatial Dropout Network.FRCN\raisebox{0.2ex}{$\star$} refers to FRCN \protect\cite{frcn} with our training schedule.}
\vspace{-0.1in}
\renewcommand{\arraystretch}{1.2}
\renewcommand{\tabcolsep}{1.2mm}
\resizebox{\linewidth}{!}{
\begin{tabular}{@{}L{3.3cm} !{\color{gray}\vrule}  L{1.2cm}  !{\color{gray}\vrule}   l  !{\color{gray}\vrule}  r*{19}{x} @{}}
\Xhline{1pt}
method  & arch & mAP & aero      & bike      & bird      & boat      & bottle     & bus        & car        & cat        & chair      & cow        & table      & dog        & horse      & mbike      & persn     & plant      & sheep      & sofa       & train      & tv   \\
\Xhline{0.8pt}

FRCN\raisebox{0.2ex}{$\star$} &AlexNet  &57.0 & 67.3 & 72.1 & 54.0 & 38.3 & 24.4 & 65.7 & 70.7 & 66.9 & 32.4 & 60.2 & 63.2 & 62.5 & 72.4 & 67.6 & 59.2 & 24.1 & 53.0 & 60.6 & 64.0 & 61.5 \\
Ours (random dropout)  &AlexNet  & 57.3 & 68.6 & 72.6 & 52.0 & 34.7 & 26.9 & 64.1 & 71.3 & 67.1 & 33.8 & 60.3 & 62.0 & 62.7 & 73.5 & 70.4 & 59.8 & 25.7 & 53.0 & 58.8 & 68.6 & 60.9 \\
Ours (hard dropout)  &AlexNet  & 57.7  & 66.3 & 72.1 & 52.8 & 32.8 & 24.3 & 66.8 & 71.7 & 69.4 & 33.4 & 61.5 & 62.0 & 63.4 & 76.5 & 69.6 & 60.6 & 24.4 & 56.5 & 59.1 & 68.5 & 62.0 \\
Ours (fixed ASDN)  &AlexNet  & 57.5 & 66.3 & 72.7 & 50.4 & 36.6 & 24.5 & 66.4 & 71.1 & 68.8 & 34.7 & 61.2 & 64.1 & 61.9 & 74.4 & 69.4 & 60.4 & 26.8 & 55.1 & 57.2 & 68.6 & 60.1 \\
Ours (joint learning) &AlexNet & 58.5 & 67.1 & 72.0 & 53.4 & 36.4 & 25.3 & 68.5 & 71.8 & 70.0 & 34.7 & 63.1 & 64.5 & 64.3 & 75.5 & 70.0 & 61.5 & 26.8 & 55.3 & 58.2 & 70.5 & 60.5 \\

\Xhline{1pt}

\end{tabular}
}
\vspace{-0.05in}
\label{tab:voc2007_ab}
\end{table*}

\subsection{Experimental settings}

\textbf{PASCAL VOC.} For the VOC datasets, we use the `trainval' set for training and `test' set for testing.  We follow most of the setup in standard Fast-RCNN~\cite{frcn} for training. We apply SGD for 80K to train our models. The learning rate starts with $0.001$ and decreases to $0.0001$ after 60K iterations. We use the selective search proposals~\cite{Uijlings13} during training. 

\textbf{MS COCO.} For the COCO dataset, we use the `trainval35k' set for training and the  `minival' set for testing. During training the Fast-RCNN~\cite{frcn}, we apply SGD with 320K iterations. The learning rate starts with $0.001$ and decreases to $0.0001$ after 280K iterations. For object proposals, we use the DeepMask proposals~\cite{DeepMask}. 

In all the experiments, our minibatch size for training is 256 proposals with 2 images. We follow the Torch implementation~\cite{Zagoruyko2016Multipath} of Fast-RCNN. With these settings, our baseline numbers for are slightly better than the reported number in~\cite{frcn}. To prevent the Fast-RCNN from overfitting to the modified data, we provide one image in the batch without any adversarial occlusions/deformations and apply our approach on another image in the batch.

\subsection{PASCAL VOC 2007 Results}
\vspace{-0.05in}
We report our results for using ASTN and ASDN during training Fast-RCNN in Table~\ref{tab:voc2007}. For the AlexNet architecture~\cite{alex}, our implemented baseline is $57.0\%$ mAP. Based on this setting, joint learning with our ASTN model reaches $58.1\%$ and joint learning with the ASDN model gives higher performance of $58.5\%$. As both methods are complementary to each other, combining ASDN and ASTN into our full model gives another boost to $58.9\%$ mAP. 

For the VGG16 architecture~\cite{VGG}, we conduct the same set of experiments. Firstly, our baseline model reaches $69.1\%$ mAP, much higher than the reported number $66.9\%$ in~\cite{frcn}. Based on this implementation, joint learning with our ASTN model gives an improvement to $69.9\%$ mAP and the ASDN model reaches $71.0\%$ mAP. Our full model with both ASTN and ASDN improves the performance to $71.4\%$. Our final result gives  $2.3\%$ boost upon the baseline. 

To show that our method also works with very deep CNNs, we apply the ResNet-101~\cite{resnet} architecture in training Fast-RCNN. As the last two lines in Table.\ref{tab:voc2007} illustrate, the performance of Fast-RCNN with ResNet-101 is $71.8\%$ mAP. By applying the adversarial training, the result is $73.6\%$ mAP. We can see that our approach consistently improves performances on different types of architectures.

\begin{table*}[t]
\centering
\caption[caption]{\small {\bf VOC 2012 test} detection average precision (\%).  FRCN\raisebox{0.2ex}{$\star$} refers to FRCN \protect\cite{frcn} with our training schedule.}
\vspace{-0.1in}

\renewcommand{\arraystretch}{1.2}
\renewcommand{\tabcolsep}{1.2mm}
\resizebox{\linewidth}{!}{

\begin{tabular}{@{}L{2.1cm} !{\color{gray}\vrule}  L{1.2cm}   l r*{19}{x} @{}}
\Xhline{1pt}
method  & arch & mAP & aero      & bike      & bird      & boat      & bottle     & bus        & car        & cat        & chair      & cow        & table      & dog        & horse      & mbike      & persn     & plant      & sheep      & sofa       & train      & tv   \\

\Xhline{0.5pt}
FRCN~\cite{frcn} &VGG & 65.7 & 80.3 & 74.7 & 66.9 & 46.9 & 37.7 & 73.9 & 68.6 & 87.7 & 41.7 & 71.1 & 51.1 & 86.0 & 77.8 & 79.8 & 69.8 & 32.1 & 65.5 & 63.8 & 76.4 & 61.7 \\
FRCN\raisebox{0.2ex}{$\star$} &VGG  & 66.4 & 81.8 & 74.4 & 66.5 & 47.8 & 39.3 & 75.9 & 69.1 & 87.4 & 44.3 & 73.2 & 54.0 & 84.9 & 79.0 & 78.0 & 72.2 & 33.1 & 68.0 & 62.4 & 76.7 & 60.8  \\
Ours (full)  &VGG  & 69.0 & 82.2 & 75.6 & 69.2 & 52.0 & 47.2 & 76.3 & 71.2 & 88.5 & 46.8 & 74.0 & 58.1 & 85.6 & 80.3 & 80.5 & 74.7 & 41.5 & 70.4 & 62.2 & 77.4 & 67.0 \\
\Xhline{1pt}
\end{tabular}
}
\vspace{-0.05in}
\label{tab:voc2012}
\end{table*}

\vspace{-0.05in}
\subsubsection{Ablative Analysis} 
\vspace{-0.05in}
\textbf{ASDN Analysis.} We compare  our Advesarial Spatial Dropout Network with various dropout/occlusion strategy in training using the AlexNet architecture. The first simple baseline we try is random spatial dropout on the feature after RoI-Pooling. For a fair comparison, we mask the activations of the same number of neurons as we do in the ASDN network. As Table~\ref{tab:voc2007_ab} shows, the performance of random dropout is $57.3\%$ mAP which is slightly better than the baseline. Another dropout strategy we compare to is a similar strategy we apply in pre-training the ASDN (Fig.~\ref{fig:ASDN}). We exhaustively enumerate different kinds of occlusion and select the best ones for training in each iteration. The performance is $57.7\%$ mAP (Ours (hard dropout)), which is slightly better than random dropout. 

As we find the exhaustive strategy can only explore very limited space of occlusion policies, we use the pre-trained ASDN network to replace it. However, when we fix the parameters of the ASDN, we find the performance is $57.5\%$ mAP (Ours (fixed ASDN) ) , which is not as good as the exhaustive strategy. The reason is the fixed ASDN has not received any feedback from the updating Fast-RCNN while the exhaustive search has.  If we jointly learn the ASDN and the Fast-RCNN together, we can get $58.5\%$ mAP, $1.5\%$ improvement compared to the baseline without dropout. This evidence shows that joint learning of ASDN and Fast-RCNN is where it makes a difference. 

\begin{figure}
    \centering
    \includegraphics[width=0.95\linewidth]{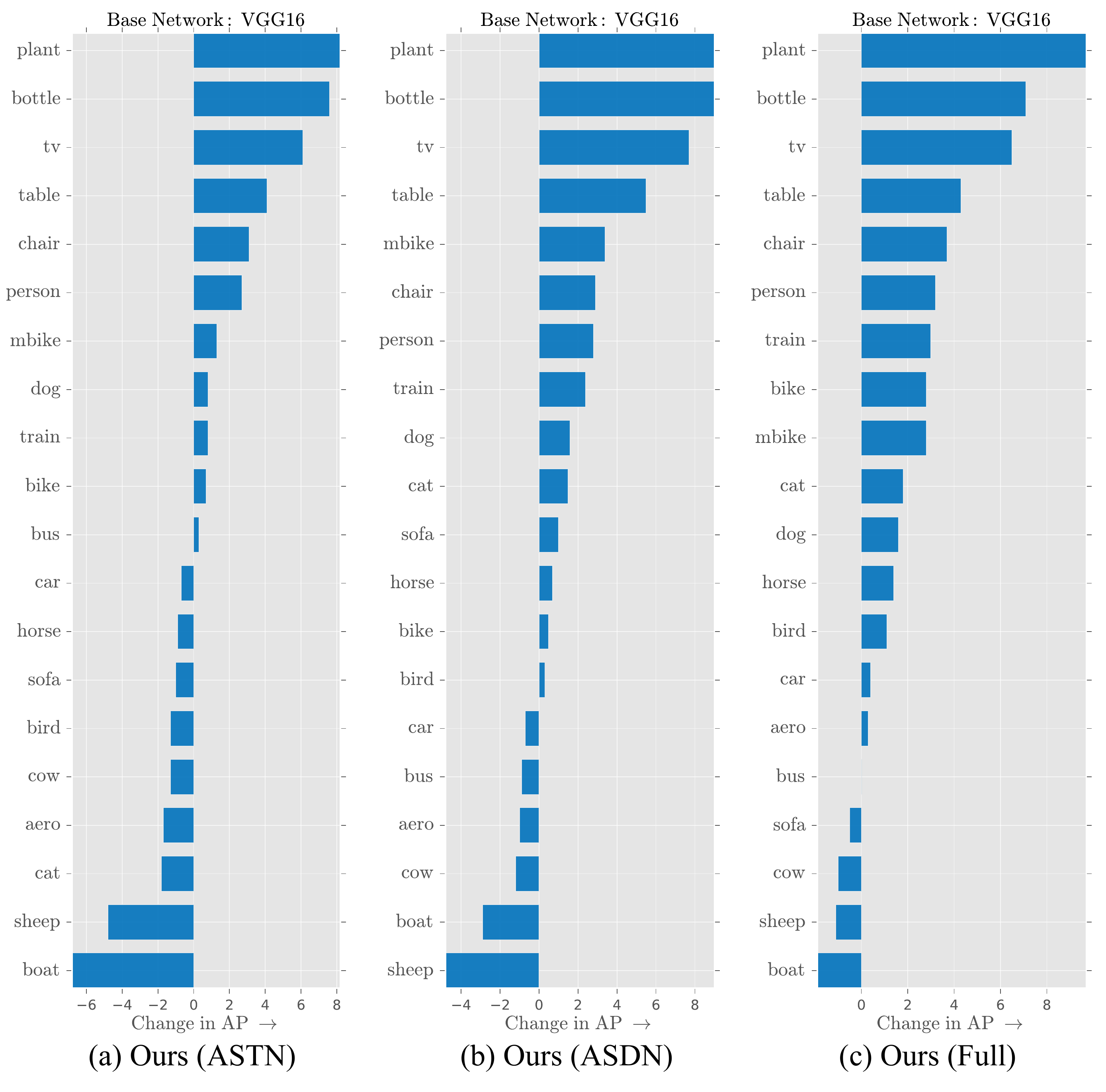}
    \vspace{-0.1in}
    \caption{Changes of APs compared to baseline FRCN.}\label{fig:exp_ap}
    \vspace{-0.1in}
\end{figure}
\textbf{ASTN Analysis.} We compared our Adversarial Spatial Transformer Network with random jittering on the object proposals. The augmentations include random changes of scale, aspect ratio and rotation on the proposals during training the Fast-RCNN. With AlexNet, the performance of using random jittering is $57.3\%$ mAP while our ASTN results is $58.1\%$. With VGG16, we have $68.6\%$ for random jittering and $69.9\%$ for the ASTN. For both architectures, the model with ASTN works better than random jittering.

\vspace{-0.1in}
\subsubsection{Category-based Analysis}
\vspace{-0.05in}
Figure~\ref{fig:exp_ap} shows the graph of how performance of each category changes with the occlusions and deformations. Interestingly the categories that seemed to be helped by both ASTN and ASDN seem to be quire similar. It seems that both {\tt plant} and {\tt bottle} performance improves with adversarial training. However, combining the two transformations together seems to improve performance on some categories which were hurt by using occlusion or deformations alone. Specifically, categories like {\tt car} and {\tt aeroplane} are helped by combining the two adversarial processes.

\subsubsection{Qualitative Results}
\begin{figure}
    \centering
    \includegraphics[width=0.95\linewidth]{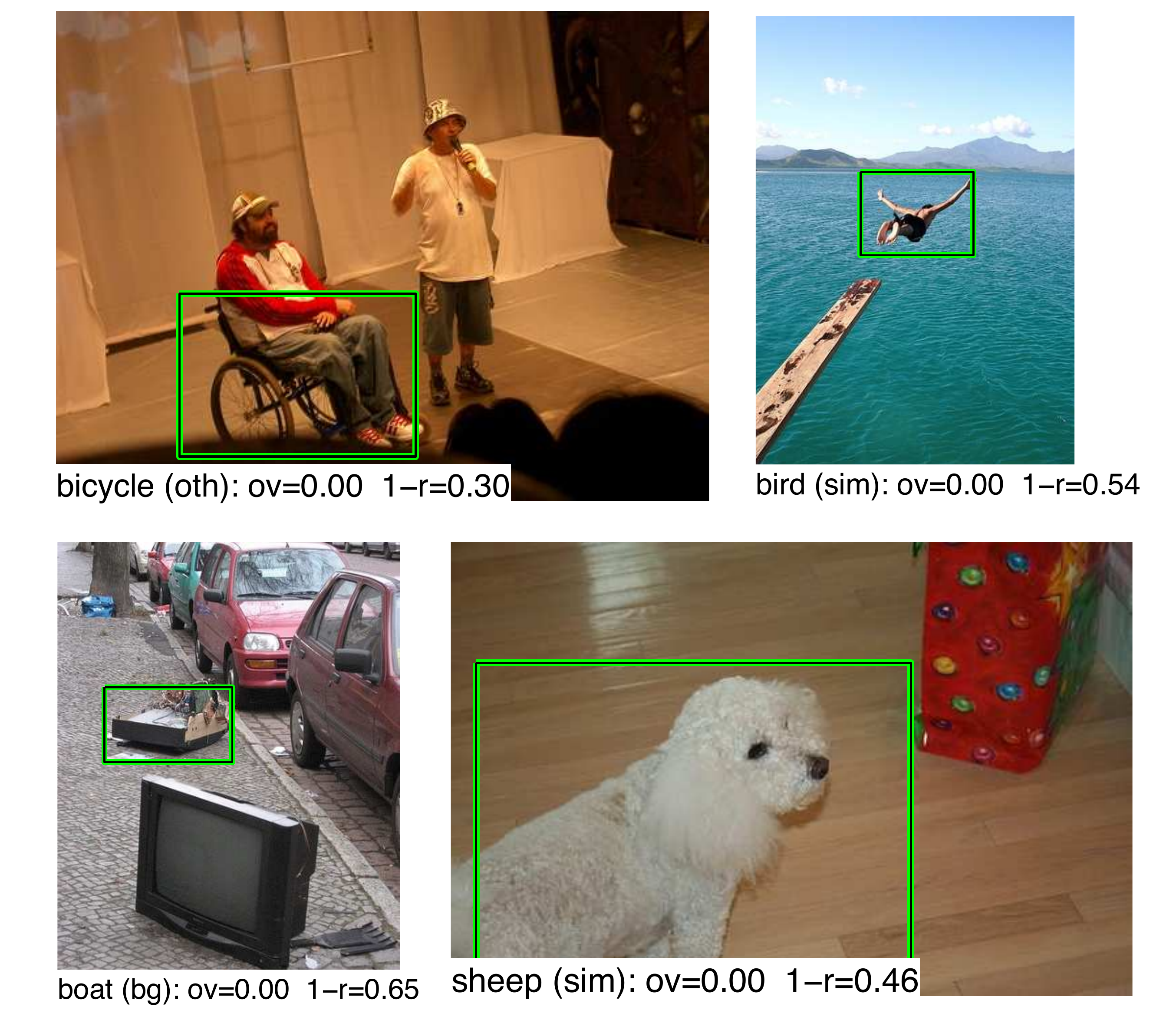}
    \vspace{-0.1in}
    \caption{Some of the false positives for our approach. These are top false positives for adversarial training but not the original Fast-RCNN.}\label{fig:exp_fp}
    \vspace{-0.15in}
\end{figure}

\vspace{-0.05in}
Figure~\ref{fig:exp_fp} shows some of the false positives of our approach with the diagnosing code~\cite{hoiem12}. These examples are hand-picked such that they only appeared in the list of false positives for adversarial learning but not the original Fast-RCNN. These results indicate some of the shortcomings of adversarial learning. In some cases, the adversary creates deformations or occlusions which are similar to other object categories and leading to over-generalization. For example, our approach hides the wheels of the bicycle which leads to a wheel chair being classified as a bike.

\subsection{Results on PASCAL VOC 2012 and MS COCO}

We show our results with VGG16 on the PASCAL VOC 2012 dataset in Table~\ref{tab:voc2012}, where our baseline performance is $66.4\%$ .Our full approach with joint learning of ASDN and ASTN gives $2.6\%$ boost to $69.0\%$ mAP. This again shows that the performance improvement using VGG on VOC2012 is significant. We also observe that our method improves performance of all the categories except sofa in VOC 2012. We believe this is probably because of larger diversity in VOC 2012.

We finally report the results in MS COCO dataset. The baseline method with VGG16 architecture is  $42.7\%$ AP$^{50}$ on the VOC metric and $25.7\%$ AP on the standard COCO metric. By applying our method, we achieve $46.2\%$ AP$^{50}$ and $27.1\%$ AP on the VOC and COCO metric respectively. 

\vspace{-0.05in}
\subsection{Comparisons with OHEM}
\vspace{-0.05in}
Our method is also related to the Online Hard Example Mining (OHEM) approach~\cite{shrivastavaOHEM}. Our method allows us to sample data-points which might not exist in the dataset, whereas OHEM is bound by the dataset. However, OHEM has more realistic features since they are extracted from real images. For comparisons, our approach ($71.4\%$) is better than OHEM ($69.9\%$) in VOC2007. However, our result ($69.0\%$) is not as good as OHEM ($69.8\%$) in VOC2012. Since these two approaches are generating or selecting different types of features in training,  we believe they should be complementary. To demonstrate this, we use an ensemble of these two approaches and compare it with separate ensembles of OHEM and Ours alone on VOC 2012. As a result, the ensemble of two methods achieves $71.7\%$ mAP, while the ensemble of two OHEM models ($71.2\%$) or two of our models ($70.2\%$) are not as good, indicating the complementary nature of two approaches.

\vspace{-0.05in}
\section{Conclusion}
\vspace{-0.05in}

One of the long-term goals of object detection is to learn object models that are invariant to occlusions and deformations. Current approaches focus on learning these invariances by using large-scale datasets. In this paper, we argue that like categories, occlusions and deformations also follow a long-tail distribution: some of them are so rare that they might be hard to sample even in a large-scale dataset. We propose to learn these invariances using adversarial learning strategy. The key idea is to learn an adversary in conjunction with original object detector. This adversary creates examples on the fly with different occlusions and deformations, such that these occlusions/deformations make it difficult for original object detector to classify. Instead of generating examples in pixel space, our adversarial network modifies the features to mimic occlusion and deformations. We show in our experiments that such an adversarial learning strategy provides significant boost in detection performance on VOC and COCO dataset.

{\footnotesize
\noindent {\bf Acknowledgement}:This work is supported by the Intelligence Advanced Research Projects Activity (IARPA) via Department of Interior/ Interior Business Center (DoI/IBC) contract number D16PC00007. The U.S. Government is authorized to reproduce and distribute reprints for Governmental purposes notwithstanding any copyright annotation thereon. Disclaimer: The views and conclusions contained herein are those of the authors and should not be interpreted as necessarily representing the official policies or endorsements, either expressed or implied, of IARPA, DoI/IBC, or the U.S. Government. AG was also supported by Sloan Fellowship.}

{\small
\bibliographystyle{ieee}
\bibliography{local}

\begin{thebibliography}{10}\itemsep=-1pt

\bibitem{ion}
S.~Bell, C.~L. Zitnick, K.~Bala, and R.~Girshick.
\newblock Inside-outside net: Detecting objects in context with skip pooling
  and recurrent neural networks.
\newblock {\em arXiv preprint arXiv:1512.04143}, 2015.

\bibitem{dai16rfcn}
J.~Dai, Y.~Li, K.~He, and J.~Sun.
\newblock {R-FCN}: Object detection via region-based fully convolutional
  networks.
\newblock {\em NIPS}, 2016.

\bibitem{imagenet}
J.~Deng, W.~Dong, R.~Socher, L.-J. Li, K.~Li, and L.~Fei-Fei.
\newblock Imagenet: A large-scale hierarchical image database.
\newblock In {\em CVPR}, 2009.

\bibitem{Denton15}
E.~Denton, S.~Chintala, A.~Szlam, and R.~Fergus.
\newblock Deep generative image models using a laplacian pyramid of adversarial
  networks.
\newblock In {\em NIPS}, 2015.

\bibitem{voc}
M.~Everingham, L.~Van~Gool, C.~K.~I. Williams, J.~Winn, and A.~Zisserman.
\newblock The pascal visual object classes (voc) challenge.
\newblock {\em IJCV}, 2010.

\bibitem{Gidaris15}
S.~Gidaris and N.~Komodakis.
\newblock Object detection via a multi- region and semantic segmentation-aware
  cnn model.
\newblock In {\em ICCV}, 2015.

\bibitem{frcn}
R.~Girshick.
\newblock Fast r-cnn.
\newblock In {\em ICCV}, 2015.

\bibitem{rcnn}
R.~Girshick, J.~Donahue, T.~Darrell, and J.~Malik.
\newblock Rich feature hierarchies for accurate object detection and semantic
  segmentation.
\newblock In {\em {CVPR}}, 2014.

\bibitem{goodfellow2014generative}
I.~Goodfellow, J.~Pouget-Abadie, M.~Mirza, B.~Xu, D.~Warde-Farley, S.~Ozair,
  A.~Courville, and Y.~Bengio.
\newblock Generative adversarial nets.
\newblock In {\em NIPS}, 2014.

\bibitem{resnet}
K.~He, X.~Zhang, S.~Ren, and J.~Sun.
\newblock Deep residual learning for image recognition.
\newblock In {\em CVPR}, 2016.

\bibitem{hoiem12}
D.~Hoiem, Y.~Chodpathumwan, and Q.~Dai.
\newblock Diagnosing error in object detectors.
\newblock In {\em ECCV}, 2012.

\bibitem{Jonathan16}
J.~Huang, V.~Rathod, C.~Sun, M.~Zhu, A.~Korattikara, A.~Fathi, I.~Fischer,
  Z.~Wojna, Y.~Song, S.~Guadarrama, and K.~Murphy.
\newblock Speed/accuracy trade-offs for modern convolutional object detectors.
\newblock In {\em CoRR}, 2016.

\bibitem{densebox}
L.~Huang, Y.~Yang, Y.~Deng, and Y.~Yu.
\newblock Densebox: Unifying landmark localization with end to end object
  detection.
\newblock In {\em CoRR}, 2015.

\bibitem{pix2pix2016}
P.~Isola, J.-Y. Zhu, T.~Zhou, and A.~A. Efros.
\newblock Image-to-image translation with conditional adversarial networks.
\newblock {\em CVPR}, 2017.

\bibitem{stn15}
M.~Jaderberg, K.~Simonyan, A.~Zisserman, and K.~Kavukcuoglu.
\newblock Spatial transformer networks.
\newblock In {\em NIPS}, 2015.

\bibitem{alex}
A.~Krizhevsky, I.~Sutskever, and G.~E. Hinton.
\newblock Imagenet classification with deep convolutional neural networks.
\newblock In {\em NIPS}, 2012.

\bibitem{FPN17}
T.~Lin, P.~Dollár, R.~Girshick, K.~He, B.~Hariharan, and S.~Belongie.
\newblock Feature pyramid networks for object detection.
\newblock In {\em CoRR}, 2017.

\bibitem{coco}
T.~Lin, M.~Maire, S.~Belongie, L.~D. Bourdev, R.~B. Girshick, J.~Hays,
  P.~Perona, D.~Ramanan, P.~Doll{\'{a}}r, and C.~L. Zitnick.
\newblock Microsoft {COCO:} common objects in context.
\newblock {\em CoRR}, abs/1405.0312, 2014.

\bibitem{ssd16}
W.~Liu, D.~Anguelov, D.~Erhan, C.~Szegedy, S.~Reed, C.-Y. Fu, and A.~C. Berg.
\newblock Ssd: Single shot multibox detector.
\newblock In {\em ECCV}, 2016.

\bibitem{loshchilov2015online}
I.~Loshchilov and F.~Hutter.
\newblock Online batch selection for faster training of neural networks.
\newblock {\em arXiv preprint arXiv:1511.06343}, 2015.

\bibitem{MathieuCL15}
M.~Mathieu, C.~Couprie, and Y.~LeCun.
\newblock Deep multi-scale video prediction beyond mean square error.
\newblock {\em CoRR}, abs/1511.05440, 2015.

\bibitem{Mirza15}
M.~Mirza and S.~Osindero.
\newblock Conditional generative adversarial nets.
\newblock {\em CoRR}, abs/1411.1784, 2014.

\bibitem{pathakCVPR16context}
D.~Pathak, P.~Kr\"ahenb\"uhl, J.~Donahue, T.~Darrell, and A.~Efros.
\newblock Context encoders: Feature learning by inpainting.
\newblock In {\em CVPR}, 2016.

\bibitem{DeepMask}
P.~O. Pinheiro, R.~Collobert, and P.~Dollár.
\newblock Learning to segment object candidates.
\newblock In {\em NIPS}, 2015.

\bibitem{pinto16}
L.~Pinto, J.~Davidson, and A.~Gupta.
\newblock Supervision via competition: Robot adversaries for learning tasks.
\newblock In {\em ICRA}, 2017.

\bibitem{Alec15}
A.~Radford, L.~Metz, and S.~Chintala.
\newblock Unsupervised representation learning with deep convolutional
  generative adversarial networks.
\newblock {\em CoRR}, abs/1511.06434, 2015.

\bibitem{yolo16}
J.~Redmon, S.~Divvala, R.~Girshick, and A.~Farhadi.
\newblock You only look once: Unified, real-time object detection.
\newblock In {\em CVPR}, 2015.

\bibitem{renNIPS15fasterrcnn}
S.~Ren, K.~He, R.~Girshick, and J.~Sun.
\newblock Faster {R-CNN}: Towards real-time object detection with region
  proposal networks.
\newblock In {\em NIPS}, 2015.

\bibitem{improvedGAN}
T.~Salimans, I.~Goodfellow, W.~Zaremba, V.~Cheung, A.~Radford, and X.~Chen.
\newblock Improved techniques for training gans.
\newblock {\em CoRR}, 2016.

\bibitem{overfeat}
P.~Sermanet, D.~Eigen, X.~Zhang, M.~Mathieu, R.~Fergus, and Y.~LeCun.
\newblock Overfeat: Integrated recognition, localization and detection using
  convolutional networks.
\newblock {\em CoRR}, abs/1312.6229, 2013.

\bibitem{ruslanattention}
S.~Sharma, R.~Kiros, and R.~Salakhutdinov.
\newblock Action recognition using visual attention.
\newblock In {\em CoRR}, 2016.

\bibitem{shrivastava16}
A.~Shrivastava and A.~Gupta.
\newblock Contextual priming and feedback for faster r-cnn.
\newblock In {\em ECCV}, 2016.

\bibitem{shrivastavaOHEM}
A.~Shrivastava, A.~Gupta, and R.~Girshick.
\newblock Training region-based object detectors with online hard example
  mining.
\newblock In {\em CVPR}, 2016.

\bibitem{TDM17}
A.~Shrivastava, R.~Sukthankar, J.~Malik, and A.~Gupta.
\newblock Beyond skip connections: Top-down modulation for object detection.
\newblock In {\em CoRR}, 2017.

\bibitem{simo2014fracking}
E.~Simo-Serra, E.~Trulls, L.~Ferraz, I.~Kokkinos, and F.~Moreno-Noguer.
\newblock Fracking deep convolutional image descriptors.
\newblock {\em arXiv preprint arXiv:1412.6537}, 2014.

\bibitem{VGG}
K.~Simonyan and A.~Zisserman.
\newblock Very deep convolutional networks for large-scale image recognition.
\newblock {\em CoRR}, abs/1409.1556, 2014.

\bibitem{CatGAN15}
J.~T. Springenberg.
\newblock Unsupervised and semi-supervised learning with categorical generative
  adversarial networks.
\newblock In {\em CoRR}, 2015.

\bibitem{inceptionresnet}
C.~Szegedy, S.~Ioffe, V.~Vanhoucke, and A.~Alemi.
\newblock Inception-v4, inception-resnet and the impact of residual connections
  on learning.
\newblock In {\em CoRR}, 2016.

\bibitem{minibatchSVM}
M.~Tak{\'a}{\v{c}}, A.~Bijral, P.~Richt{\'a}rik, and N.~Srebro.
\newblock Mini-batch primal and dual methods for svms.
\newblock {\em arXiv preprint arXiv:1303.2314}, 2013.

\bibitem{Uijlings13}
J.~Uijlings, K.~van~de Sande, T.~Gevers, and A.~Smeulders.
\newblock Selective search for object recognition.
\newblock {\em International Journal of Computer Vision}, 2013.

\bibitem{wang2015unsupervised}
X.~Wang and A.~Gupta.
\newblock Unsupervised learning of visual representations using videos.
\newblock In {\em ICCV}, 2015.

\bibitem{reinforce}
R.~J. Williams.
\newblock Simple statistical gradient-following algo- rithms for connectionist
  reinforcement learning.
\newblock In {\em Machine learning}, 1992.

\bibitem{Xie2016}
S.~Xie, R.~Girshick, P.~Dollár, Z.~Tu, and K.~He.
\newblock Aggregated residual transformations for deep neural networks.
\newblock {\em arXiv preprint arXiv:1611.05431}, 2016.

\bibitem{Zagoruyko2016Multipath}
S.~Zagoruyko, A.~Lerer, T.-Y. Lin, P.~O. Pinheiro, S.~Gross, S.~Chintala, and
  P.~Doll{\'{a}}r.
\newblock A multipath network for object detection.
\newblock In {\em BMVC}, 2016.

\bibitem{Zengeccv16}
X.~Zeng, W.~Ouyang, B.~Yang, J.~Yan, and X.~Wang.
\newblock Gated bi-directional cnn for object detection.
\newblock {\em ECCV}, 2016.

\end{thebibliography}
}

\end{document}